\documentclass[iicol]{sn-jnl}

\usepackage{amssymb}
\usepackage{graphicx}
\usepackage{amsmath}
\usepackage{booktabs}
\usepackage{array}
\usepackage{tabularx}
\usepackage{multirow}
\usepackage{lipsum}
\usepackage{threeparttable}
\usepackage{setspace}
\usepackage{lmodern}
\usepackage{orcidlink}
\usepackage{xcolor}
\usepackage{ulem}
\usepackage{mathrsfs}
\usepackage{adjustbox}
\usepackage{stfloats}
\usepackage[switch]{lineno}

\newcolumntype{Y}{>{\centering\arraybackslash}X}

\title[mode=title]{Deep Learning for Detecting and Early Predicting Chronic Obstructive Pulmonary Disease from Spirogram Time Series}

\author[1,7]{\fnm{Shuhao}
\sur{Mei}~\orcidlink{0009-0007-9505-3699}}
\equalcont{These authors contributed equally to this work.}

\author[9]{\fnm{Xin}
\sur{Li}}
\equalcont{These authors contributed equally to this work.}

\author*[1,2]{\fnm{Yuxi} \sur{Zhou}}\email{joy\_yuxi@pku.edu.cn}

\author[1]{\fnm{Jiahao}
\sur{Xu}}

\author*[2]{\fnm{Yong} \sur{Zhang}}\email{zhangyong05@tsinghua.edu.cn}

\author[1]{\fnm{Yuxuan}
\sur{Wan}}

\author[3]{\fnm{Shan}
\sur{Cao}}

\author[4]{\fnm{Qinghao}
\sur{Zhao}}

\author[5]{\fnm{Shijia}
\sur{Geng}}

\author[6]{\fnm{Junqing}
\sur{Xie}}

\author[1]{\fnm{Shengyong}
\sur{Chen}}

\author*[7,8]{\fnm{Shenda}
\sur{Hong}~\orcidlink{0000-0001-7521-5127}}\email{hongshenda@pku.edu.cn}

\affil[1]{\orgdiv{Department of Computer Science},
            \orgname{Tianjin University of Technology}, 
            \orgaddress{\city{Tianjin},\country{China}}}
            
\affil[2]{\orgdiv{DCST, BNRist, RIIT, Institute of Internet Industry},
            \orgname{Tsinghua University}, 
            \orgaddress{\city{Beijing},
            \country{China}}}
\affil[3]{\orgdiv{Department of Biotheraphy}, 
            \orgname{The Second Hospital of Tianjin Medical University},
            \orgaddress{\city{Tianjin},
            \country{China}}}
\affil[4]{\orgdiv{Department of Cardiology},
            \orgname{Peking University People’s Hospital},
            \orgaddress{\city{Beijing},
            \country{China}}}
\affil[5]{\orgdiv{HeartVoice Medical Technology},
            \orgaddress{\city{Hefei},
            \country{China}}}
\affil[6]{\orgdiv{Centre for Statistics in Medicine and NIHR Biomedical Research Centre Oxford},
            \orgname{University of Oxford}, 
            \orgaddress{\city{Oxford},
            \country{UK}}}
\affil[7]{\orgdiv{National Institute of Health Data Science},
            \orgname{Peking University},
            \orgaddress{\city{Beijing},
            \country{China}}}
\affil[8]{\orgdiv{Institute for Artificial Intelligence},
            \orgname{Peking University}, 
            \orgaddress{\city{Beijing},
            \country{China}}}
\affil[9]{\orgdiv{Department of Rehabilitation Medicine, Beijing Tsinghua Changgung Hospital, School of Clinical Medicine},
                \orgname{Tsinghua University},
                \orgaddress{\city{Beijing},
                \country{China}}}

\abstract{Chronic Obstructive Pulmonary Disease (COPD) is a chronic lung condition characterized by airflow obstruction. Current diagnostic methods primarily rely on identifying prominent features in spirometry (Volume-Flow time series) to detect COPD, but they are not adept at predicting future COPD risk based on subtle data patterns. In this study, we introduce a novel deep learning-based approach, DeepSpiro, aimed at the early prediction of future COPD risk. DeepSpiro consists of four key components: SpiroSmoother for stabilizing the Volume-Flow curve, SpiroEncoder for capturing volume variability-pattern through key patches of varying lengths, SpiroExplainer for integrating heterogeneous data and explaining predictions through volume attention, and SpiroPredictor for predicting the disease risk of undiagnosed high-risk patients based on key patch concavity, with prediction horizons of 1, 2, 3, 4, 5 years, or even longer. Evaluated on the UK Biobank dataset, DeepSpiro achieved an AUC of 0.8328 for COPD detection and demonstrated strong predictive performance for future COPD risk (p-value \textless 0.001). In summary, DeepSpiro can effectively predicts the long-term progression of the COPD disease.
}

\begin{document}
\maketitle

\section{Introduction}\label{Introduction}
Chronic Obstructive Pulmonary Disease (COPD) is a progressively worsening lung disease that leads to difficulty breathing, limited activity, and a decline in quality of life \cite{agusti2023global,ferrera2021advances}. As the disease progresses, COPD may also increase the risk of cardiovascular diseases \cite{maeda2024chronic} and even lead to premature death. Therefore, timely and accurate COPD detection is crucial to reduce patient health risks \cite{stolz2022towards}. Previous studies have shown a strong correlation between detecting the disease at an early stage and the success of its treatment. Failing to identify the disease during this crucial period will worsen its severity \cite{aaron2024early,macleod2021chronic}.

Clinical diagnosis often involves identifying COPD patients by determining if the FEV1/FVC ratio is below 70\% \cite{toren2021ratio,bhatt2019discriminative,hoesein2011lower}. 
Researchers have discovered that the FEV1/FVC ratio method isn't always accurate when used with people of different ages \cite{bhatt2023fev1,medbo2007lung}. This means that some patients might not get the right diagnosis at the right time, missing out on early and personalized treatment options.

In recent years, researchers have used deep learning technologies \cite{cosentino2023inference} to identify COPD characteristics by analyzing the Volume-Flow curve. 
This method helps to overcome the limitations of traditional approaches but cannot effectively predict an individual's latent risk of developing COPD. Moreover, introducing deep learning technology has resulted in models lacking transparency, making it difficult to gain the trust of medical professionals and patients \cite{fernandez2023deep}. Therefore, developing an artificial intelligence interpretable algorithm that can accurately detect patients with COPD and early predict an individual's latent risk of COPD is crucial for slowing disease progression and preventing patient mortality. 

\begin{figure}[]
\centering
\includegraphics[width=0.5\textwidth]{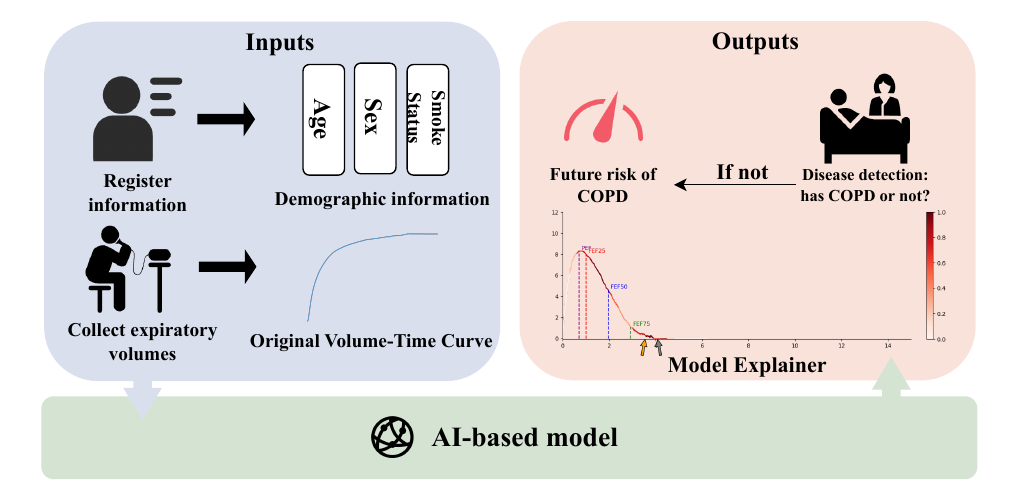}
\caption{Our input module uses the raw Time-Volume curve time series collected from hospitals and patient demographic data, which are then passed into our AI-based model module. The AI-based model module is divided into four tasks (see Section \ref{overview} for details). After processing through the AI-based model module, the output data is handled by the output module. If the AI-based model diagnoses the individual as a COPD patient, we will output their diagnosis results and the interpretability figure of the model. If the AI-based model diagnoses the individual as a non-COPD patient, we will output their risk of developing COPD over the next 1-5 years.}
\label{input-output}
\end{figure}

However, most existing methods for COPD risk analysis mainly face the following four challenges:
\begin{itemize}
\item In order to determine the degree of airflow obstruction, it is necessary to generate Volume-Flow curves based on the original Time-Volume curves. Nevertheless, current techniques for generating Volume-Flow curves can yield unstable curves, potentially resulting in incorrect predictions by the model.
\item The length of the Volume-Flow curve varies among individuals due to differences in exhalation durations. Nevertheless, existing techniques commonly handle Volume-Flow curves with varying lengths by either filling in missing values or truncating through downsampling. The former can readily add extraneous noise, whereas the latter could sacrifice important data dependencies.
\item Existing deep learning models are still black-box models that can only produce detection outcomes without offering an explainer for those results. Models lacking transparency may find it challenging to gain the trust of medical professionals and patients.
\item Currently, existing methods can only detect patients who have already had COPD based on obvious characteristics displayed on the spirogram (In this article, the spirogram specifically involves measuring Volume-Flow curve time series). However, these methods fail to early predict an individual's probability of COPD in the future based on changes in the spirogram.
\end{itemize}

To address the aforementioned challenges, we propose DeepSpiro, a method based on deep learning for early prediction of future COPD risk (Figure \ref{input-output}). Specifically, this paper makes four major contributions:

\begin{itemize}
\item We apply a method for constructing Volume-Flow curves guided by Time-Volume instability smoothing (SpiroSmoother), which uses a curve smoothing algorithm to precisely enhance the stability of the Volume-Flow curve while retaining the essential physiological information from the original Volume-Flow data.

\item We develop a COPD identification method based on learning from varied-length key patch variability-pattern (SpiroEncoder). This algorithm can dynamically calculate the "key patch" number that is best suited for each time series data patch, thereby unifying the time series representation and extracting key physiological information from the original high-dimensional dynamic sequence into a unified low-dimensional time series representation.

\item We propose a method for explaining the model based on volume attention and heterogeneous feature fusion (SpiroExplainer). This method combines the probability of having COPD with demographic data like age, sex, and probability. This information is then used as input for the model, which outputs a COPD risk assessment and offers an explainer for the model's decisions regarding the individual.

\item We develop a method for predicting the risk of COPD based on the variability-pattern of key patch concavity (SpiroPredictor) for the first time. Our method can precisely forecast the probability of disease onset in undiagnosed high-risk patients over the next 1, 2, 3, 4, 5 years, and beyond. Additionally, it can accurately categorize these patients, thereby addressing the current deficiency in early predicting future COPD risk.
\end{itemize}

Our work has made COPD detection and early risk prediction more accurate, ultimately contributing to improved clinical decision-making and patient prognosis. By providing interpretable results and predicting future risks, DeepSpiro has the potential to become a valuable early screening tool. This could help delay disease progression and may potentially reduce patient mortality.

\section{Results}

\begin{figure*}[]
\includegraphics[width=1\textwidth]{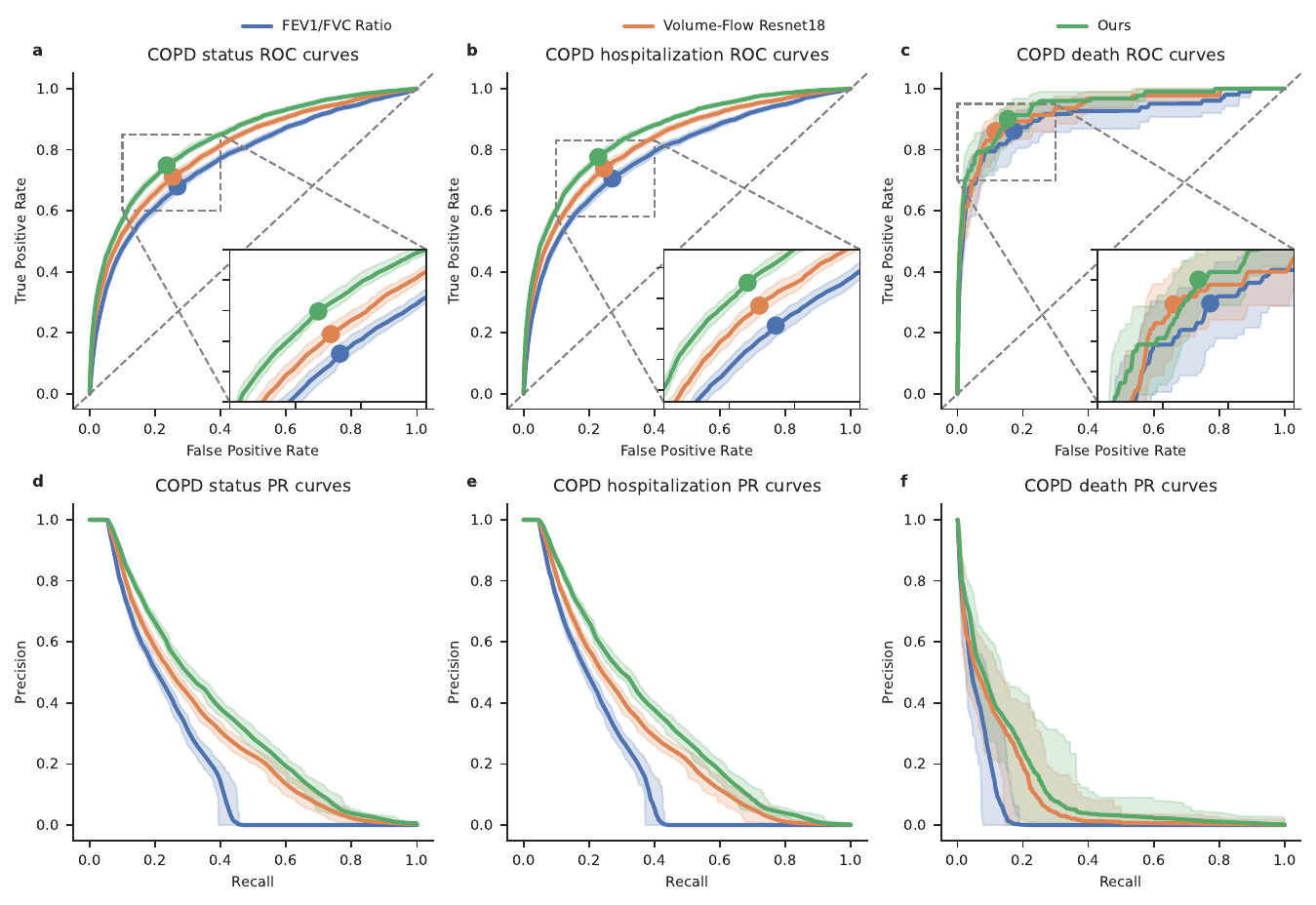}
\centering
\caption{Evaluation comparison. We compared the performance of three methods for detecting COPD: the FEV1/FVC ratio (with a threshold of 0.7), the ResNet18 model, and the DeepSpiro model. The evaluation metrics, AUROC and AUPRC, were assessed on three datasets: the full dataset, the hospitalization dataset (Only includes individuals registered as inpatients), and the death dataset (Only includes individuals registered as deceased).}
\label{evaluation-comparison-fig}
\end{figure*}

\subsection{Model Detection and Prediction Performance}
In this study, we aim to detect COPD and validate DeepSpiro superiority by comparing it to a ResNet18 baseline model. Previously, Justin Cosentino et al. proposed a baseline model in a paper published in Nature Genetics \cite{cosentino2023inference}. It was trained and evaluated on the UK Biobank dataset, with evaluation metrics including AUROC, AUPRC, and F1-score. To ensure a fair comparison, we used the same dataset and evaluation criteria. In the COPD detection task, DeepSpiro achieved an AUROC of 83.28\%, an AUPRC of 35.70\%, and an F1-score of 39.50\% as shown in Figure \ref{evaluation-comparison-fig}. DeepSpiro outperformed the baseline model in these metrics by 1.16\%, 1.24\%, and 1.36\%, respectively.

Table \ref{evaluation-comparison} presents a comparison of different metrics prior to and following signal enhancement. Based on the experimental findings, our signal enhancement technique effectively improves the stability of the Volume-Flow curve without compromising the essential physiological information of the original Volume-Flow data. This enhancement greatly aids in evaluating patients with COPD.

DeepSpiro emphasizes data integrity and the preservation of key information, in contrast to the Volume-Flow ResNet18 method (Baseline model). Based on the experimental findings, DeepSpiro betters the Volume-Flow ResNet18 method in all categories, as evidenced by higher values of key metrics such as AUROC, AUPRC, and F1-score.

\begin{figure}[]
\centerline{\includegraphics[width=0.5\textwidth]{nomogram.pdf}}
\caption{The nomogram of COPD detection. The nomogram illustrates the contribution of demographic information and the FEV1/FVC diagnostic gold standard to the model's diagnostic accuracy for COPD. The nomogram allows for visual estimation of the probability of COPD diagnosis by assigning weighted scores to each variable. The “threshold” in the figure represents the cutoff value at which the predicted probability indicates a positive diagnosis for COPD. For instance, a threshold of 0.5 means that a predicted probability greater than 0.5 would be considered indicative of COPD.}
\label{nomogram-fig}
\end{figure}

In terms of parameters, DeepSpiro also performs exceptionally well. In order to demonstrate the specific advantages of our proposed DeepSpiro with respect to certain parameters, a comparative study is carried out between DeepSpiro and the Volume-Flow ResNet18 model. Specifically, Pytorch is used to build both models and utilize the thop library to calculate both models' parameter count and floating-point operations. The calculation results are shown in Supplementary Table 4. The results indicate that DeepSpiro's parameter count and floating-point operations are significantly lower than those of the Volume-Flow ResNet18 model.

By leveraging demographic information, the model can gain a more comprehensive understanding of the patient's background \cite{sun2024covid}, and on this basis, improve its prediction accuracy. Additionally, incorporating FEV1/FVC, the gold standard for COPD diagnosis, into the model can enhance the model's capability in diagnosing COPD. To more comprehensively reflect the improvement in the model's diagnostic capability by demographic information and the FEV1/FVC diagnostic gold standard \cite{mirza2018copd,han2021gold,venkatesan2024gold}, we concatenate the probabilities outputted by module II (Figure \ref{framwork-overview-fig}-II), demographic information, and the FEV1/FVC diagnostic gold standard into a new vector and input it into a logistic regression model. Figure \ref{nomogram-fig} is a nomogram \cite{hong2023simplenomo}, from which we can intuitively see that the added demographic information and the FEV1/FVC diagnostic gold standard enhance the model's diagnostic accuracy. Therefore, the model explainer method based on volume attention and heterogeneous feature fusion has a promotive effect on the diagnosis of COPD.

\begin{table*}[ht]
\centering
\caption{Evaluation comparison table. This table presents the AUROC, AUPRC, and F1 scores for detecting COPD using four different methods: the FEV1/FVC method, the ResNet18 model, DeepSpiro trained on unsmoothed curves, and DeepSpiro trained on smoothed curves.}
\label{evaluation-comparison}
\begin{tabularx}{\textwidth}{llYYY}
\toprule
Category & Method & AUROC & AUPRC & F1-Score \\
\midrule
\multirow{4}{*}{All} & FEV1/FVC \cite{mirsadraee2015evaluation} & 0.7771 & 0.2266 & 0.3143 \\
                     & Volume-Flow ResNet18 \cite{cosentino2023inference} & 0.8212 & 0.3446 & 0.3814 \\
                     & DeepSpiro (non-smooth) & 0.8315 & 0.3537 & 0.3913 \\
                     & DeepSpiro & 0.8328 & 0.3570 & 0.3950 \\
\midrule
\multirow{4}{*}{Hospitalization} & FEV1/FVC \cite{mirsadraee2015evaluation} & 0.7925 & 0.2079 & 0.3024 \\
                                  & Volume-Flow ResNet18 \cite{cosentino2023inference} & 0.8382 & 0.3247 & 0.3703 \\
                                  & DeepSpiro (non-smooth) & 0.8527 & 0.3453 & 0.3871 \\
                                  & DeepSpiro & 0.8538 & 0.3490 & 0.3900 \\
\midrule
\multirow{4}{*}{Death} & FEV1/FVC \cite{mirsadraee2015evaluation} & 0.9242 & 0.0393 & 0.1206 \\
                        & Volume-Flow ResNet18 \cite{cosentino2023inference} & 0.9374 & 0.0887 & 0.2026 \\
                        & DeepSpiro (non-smooth) & 0.9419 & 0.1129 & 0.2211 \\
                        & DeepSpiro & 0.9419 & 0.1165 & 0.2304 \\
\bottomrule
\end{tabularx}
\end{table*}

In summary, traditional pulmonary function assessment indicators cannot fully reflect the disease's complexity and patients' overall health status. As shown in Table \ref{evaluation-comparison}, DeepSpiro is more advanced than traditional pulmonary function assessment indicators and the Volume-Flow ResNet18 model.

\subsection{Future Risk Prediction Analysis}
\begin{figure*}[]
\includegraphics[width=1\textwidth]{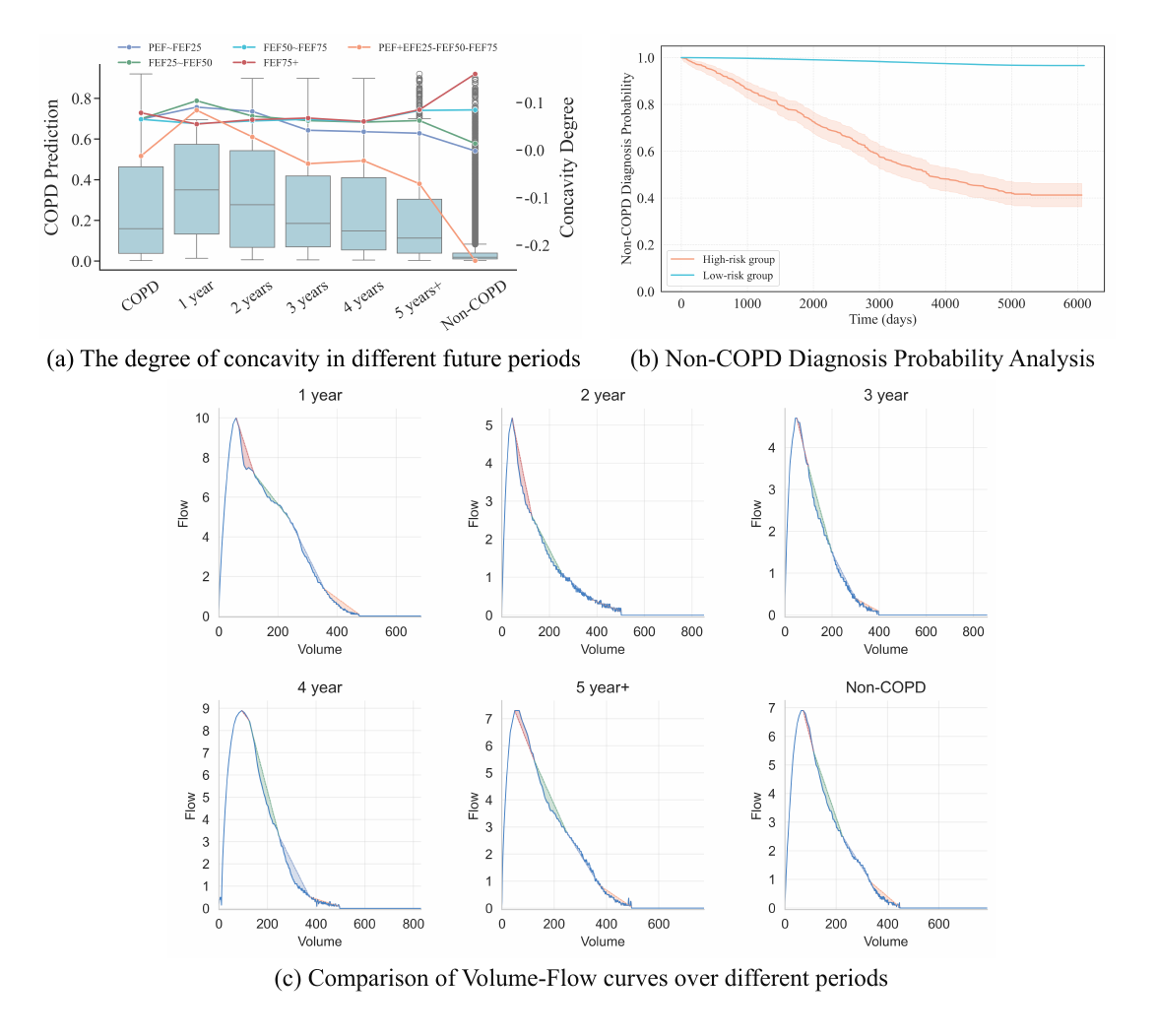}
\centering
\caption{(a) The left vertical axis represents the predicted probability of COPD, while the right vertical axis indicates the concavity degree based on the directed area metric for each phase. Each plot displays the mean concavity degree for the respective phase, illustrating how the concavity changes over time in relation to COPD risk. (b) This figure illustrates the probabilities of not being diagnosed with COPD over time for the high-risk and low-risk groups, as predicted by the model. The X-axis represents the time since the pulmonary function test, while the Y-axis shows the probability of not being diagnosed with COPD at each time point. Due to right censoring, not all high-risk patients are diagnosed within the observation period, resulting in probabilities that remain above zero. (c) As the onset time progresses, the concavity of the patient’s Volume-Flow curve decreases year by year.}
\label{merged1}
\end{figure*}

To better explain the results, we categorize the curves into two groups: early-phase curves (PEF$\sim$FEF25 and FEF25$\sim$FEF50) and late-phase curves (FEF50$\sim$FEF75 and FEF75+). As shown in Figure \ref{merged1}-(a), by observing the trend of changes in the two types of curves, we can see a significant difference in the degree of concavity. The concavity of the early-phase curves decreases as disease risk decreases, while the concavity of the late-phase curves increases as disease risk decreases. By observing Figure \ref{merged1}-(a), we find that individuals with higher disease risk (e.g., 1 year) are more likely to experience curve collapse in the early phases, whereas individuals with lower risk (e.g., Non-COPD) tend to exhibit curve collapse in the late phases. As the risk of disease increases, the phase in which the curve collapse occurs tends to shift earlier.

To more intuitively demonstrate the relationship between disease risk and the timing of curve collapse, we designed a concavity trend measurement model. This model uses the formula: $\mathscr{C}(PEF\sim FEF25)$ + $\mathscr{C}(FEF25\sim FEF50)$ - $\mathscr{C}(FEF50\sim FEF75)$ - $\mathscr{C}(FEF75+)$ to measure the concavity trend of the curve. From the formula, it can be seen that a larger concavity trend value indicates that the curve collapse (the phase where expiratory flow rate drops most significantly) occurs in the earlier phases, whereas a smaller concavity trend value indicates that the curve collapse occurs in the later phases.

Figure \ref{merged1}-(b) includes the future disease risk of high-risk and low-risk populations. The model predicts individuals as a high-risk group for those identified as at risk of disease and as a low-risk group for those predicted not to have the disease. We plotted the changes in disease risk for both groups as curves, showing the proportion diagnosed with COPD over time. The X-axis represents the time elapsed since the pulmonary function test, and the Y-axis represents the proportion of patients diagnosed with COPD at a specific timestamp within the population, i.e., the risk of disease. From the figure, it can be seen that from the time of the pulmonary function test, the disease risk for the model-predicted high-risk group shows an exponential increase over time, while the disease risk for the low-risk group remains near zero, essentially unchanged with a slight and weak increase. High-risk and low-risk groups also exhibit significant differences, with a p-value of \textless0.001.

In order to obtain the expected Volume-Flow curve for different periods, we used the k-median algorithm to extract features from the model for samples corresponding to different risk periods (1 year, 2 years, 3 years, 4 years, 5+ years, Non-COPD). We then determined the median sample for each risk period and used it as the expected Volume-Flow curve for that stage. Similarly, to obtain the expected Volume-Flow curve for different subgroups, we subdivided the groups according to their classifications and used k-median analysis to find the median sample for each subgroup, using it to represent the expected Volume-Flow curve for that subgroup.

As shown in Figure \ref{merged1}-(c), as the onset time progresses, the concavity of the patient’s Volume-Flow curve decreases year by year. This trend also indicates that the impact of the disease on lung function is more significant in the later stages, and the closer it is to the onset time, the more apparent the disease becomes.

From the predictive distribution probabilities over future times and the future disease risk scenarios for high-risk and low-risk populations, it can be seen that DeepSpiro can effectively predict the future development trends of the disease, thereby providing better treatment and management plans for patients.

\subsection{Result of Subgroup Analysis}
To further explore the diagnostic accuracy of the model across different subsets of the population, subgroup analyses for various age groups, sex, and smoking statuses are conducted.

\begin{figure*}[]
\includegraphics[width=1\textwidth]{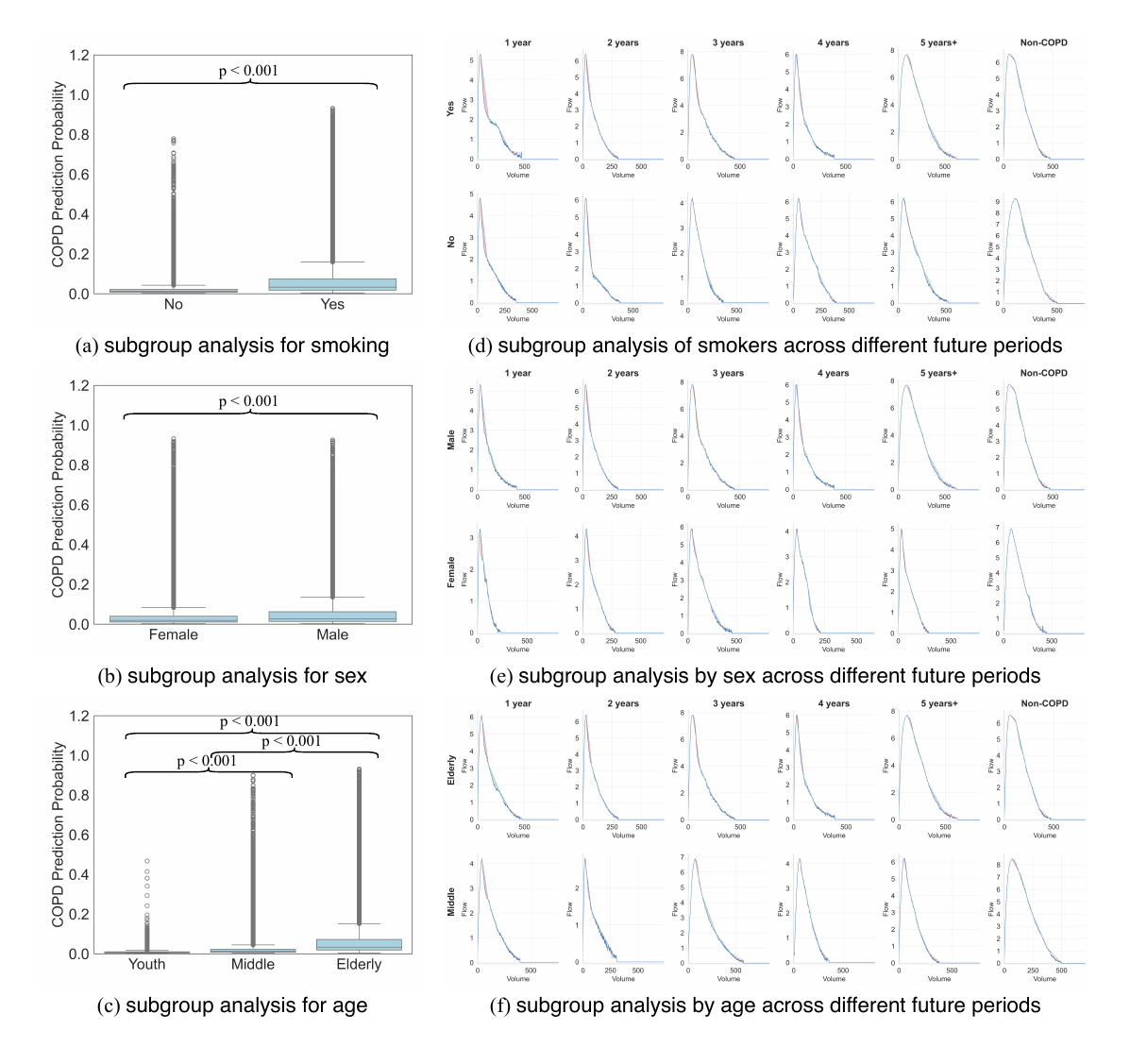}
\centering
\caption{(a) The subgroup analysis for smoking. (b) The subgroup analysis by sex. (c) The subgroup analysis by age. (d) As the onset time progresses, an individual’s concavity measure gradually decreases. Compared to non-smokers, smokers show significantly higher lung function concavity measures. (e) As the onset time progresses, an individual’s concavity measure gradually decreases. Compared to females, males show significantly higher lung function concavity measures. (f) As the onset time progresses, an individual’s concavity measure gradually decreases. Compared to younger patients, older patients show significantly higher lung function concavity measures.}
\label{merged2}
\end{figure*}

The subgroup analysis results for smokers illustrate the model's application scenarios. Among patients with COPD, the prevalence rate in smokers is higher than in non-smokers \cite{yang2022chronic}. For our model, as shown in Figure \ref{merged2}-(a), the prediction probability for smokers is higher than for non-smokers, and the prediction range for smokers is also wider than that for the non-smoking population. This is consistent with clinical significance. Regarding the p-value, the model demonstrates a strong discriminative ability between the two groups.

\begin{table}[ht]
\centering
\caption{Comparison of subgroup evaluations. This table shows the performance of different methods in detecting COPD across various subgroup analyses. These methods include the FEV1/FVC method, the ResNet18 model, and DeepSpiro trained on smoothed Volume-Flow curves.}
\label{evaluation_comparison_subgroup}
\begin{tabularx}{\columnwidth}{lllY}
\toprule
Group & Class & Model & AUROC \\
\midrule
\multirow{3}{*}{Sex} 
 & Male & FEV1/FVC \cite{mirsadraee2015evaluation} & 0.7719 \\
 &      & ResNet18 \cite{cosentino2023inference} & 0.8166 \\
 &      & DeepSpiro & 0.8335 \\
\cmidrule(l){2-4}
 & Female & FEV1/FVC \cite{mirsadraee2015evaluation} & 0.7629 \\
 &        & ResNet18 \cite{cosentino2023inference} & 0.8196 \\
 &        & DeepSpiro & 0.8329 \\
\midrule
\multirow{3}{*}{Smoke} 
 & Yes & FEV1/FVC \cite{mirsadraee2015evaluation} & 0.7774 \\
 &     & ResNet18 \cite{cosentino2023inference} & 0.8223 \\
 &     & DeepSpiro & 0.8266 \\
\cmidrule(l){2-4}
 & No & FEV1/FVC \cite{mirsadraee2015evaluation} & 0.6857 \\
 &    & ResNet18 \cite{cosentino2023inference} & 0.7428 \\
 &    & DeepSpiro & 0.7620 \\
\midrule
\multirow{3}{*}{Age} 
 & Youth & FEV1/FVC \cite{mirsadraee2015evaluation} & 0.6960 \\
 &       & ResNet18 \cite{cosentino2023inference} & 0.7344 \\
 &       & DeepSpiro & 0.7634 \\
\cmidrule(l){2-4}
 & Middle & FEV1/FVC \cite{mirsadraee2015evaluation} & 0.7325 \\
 &             & ResNet18 \cite{cosentino2023inference} & 0.7779 \\
 &             & DeepSpiro & 0.8022 \\
\cmidrule(l){2-4}
 & Elderly & FEV1/FVC \cite{mirsadraee2015evaluation} & 0.7622 \\
 &         & ResNet18 \cite{cosentino2023inference} & 0.8024 \\
 &         & DeepSpiro & 0.8210 \\
\bottomrule
\multicolumn{4}{l}{\footnotesize Note: ResNet18 in the table refers to Volume-Flow ResNet18.}
\end{tabularx}
\end{table}

The subgroup analysis results for sex showcase the model's application scenarios. The 2018 China Adult Lung Health Study, which has surveyed 50,991 individuals across ten provinces and cities, has shown that the incidence rate among men is higher than that among women, with men at 11.9\% and women at 5.4\%. For our model, as shown in Figure \ref{merged2}-(b), the prediction probability for men is higher than for women, and the prediction range for men is also greater than that for women. This aligns with clinical significance. Regarding the p-value, the model demonstrates a strong discriminative ability between the two groups.

Age is a significant factor influencing the development of COPD. Typically, the high-incidence age range for COPD is between 45-80 years. Within this age range, patients' lung functions tend to decline gradually. Particularly after age 45, the trend of declining lung function becomes more apparent. After reaching 80 years, lung function further declines, and patients' immune systems weaken, making them relatively more susceptible to diseases. Therefore, prevention and early identification of COPD are especially important for this age group. We divide patients into Youth (18-44), Middle (45-55), and Elderly (55 and above) for subgroup analysis, as shown in Figure \ref{merged2}-(c). The results indicate that DeepSpiro's prediction probability for smokers is higher than for non-smokers, and the prediction range for smokers is also larger than for the non-smoking population, consistent with clinical significance. Regarding the p-value, the model shows strong discriminative ability between the two groups.

Additionally, we compare the AUROC of DeepSpiro, the Volume-Flow ResNet18 model, and the FEV1/FVC metric across different subgroups, as shown in Table \ref{evaluation_comparison_subgroup}. Across all subgroups, DeepSpiro's AUROC is the highest, demonstrating DeepSpiro's applicability in different subgroups.

We also conducted a subgroup analysis for individuals across different future periods. As shown in Figure \ref{merged2}-(d), as the onset time progresses, an individual’s concavity measure gradually decreases. Compared to non-smokers, smokers show significantly higher lung function concavity measures, further confirming the impact of smoking on COPD. This trend is consistent with clinical observations, highlighting the association between long-term smoking and lung function decline.

As shown in Figure \ref{merged2}-(e), as the onset time progresses, an individual’s concavity measure gradually decreases. Compared to females, males show significantly higher lung function concavity measures, which is typically related to higher smoking rates, different lifestyle habits, and physiological differences in men. This aligns with the expected results from clinical research.

As shown in Figure \ref{merged2}-(f), as the onset time progresses, an individual’s concavity measure gradually decreases. Compared to younger patients, older patients show significantly higher lung function concavity measures, which may be related to natural lung function decline with age, weakened immunity, and other factors. This is consistent with clinical observations and expected results. (Note: Due to the small sample size in the Youth (18-44) group, it was not possible to subdivide and compare across different stages, so only the Middle (45-55) and Elderly (55+) groups were compared.)

\subsection{Result of SHAP Analysis}
To further analyze whether the model's predictions hold clinical significance, we conduct a SHAP analysis \cite{alabdullah2022prediction}.

\begin{figure*}[]
\includegraphics[width=1\textwidth]{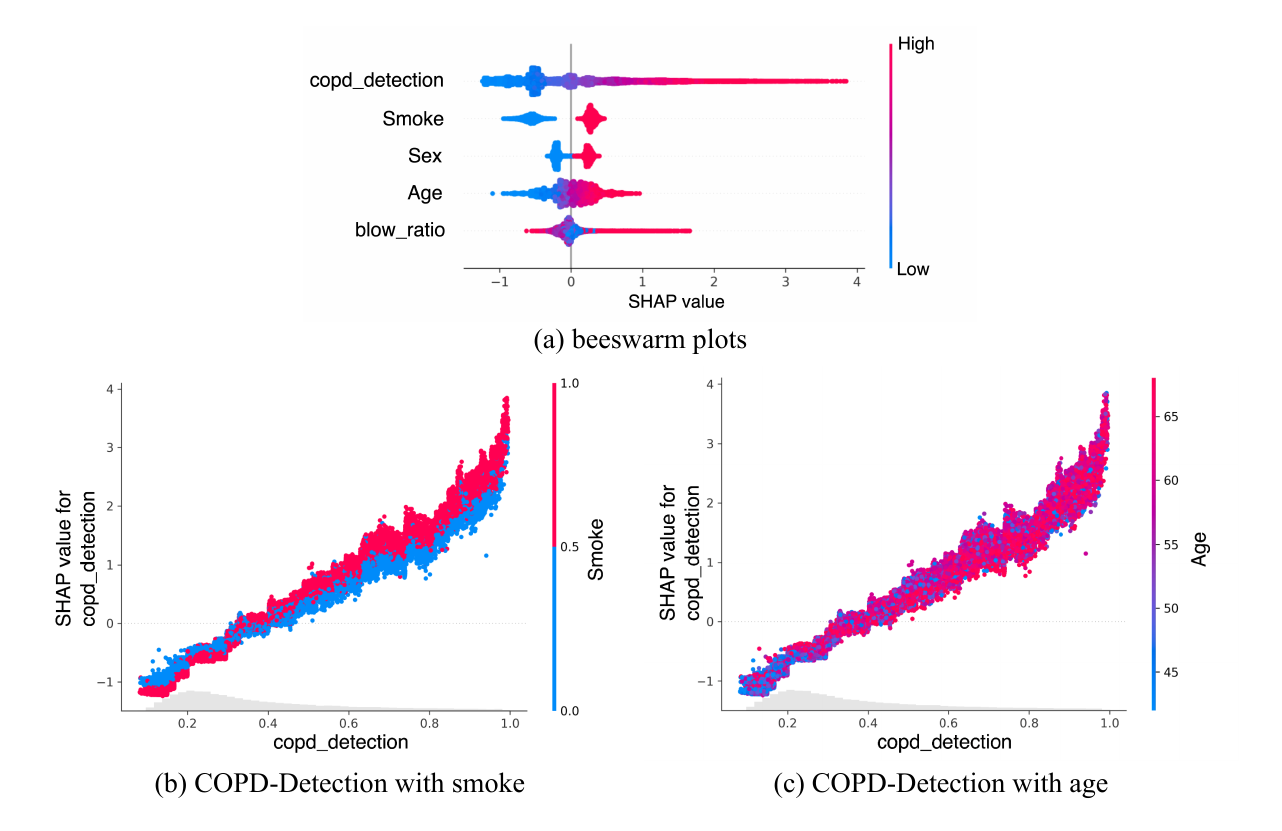}
\centering
\caption{(a) Brighter colors mean that the feature has a more positive impact on the model's predictions. The blow ratio represents the FEV1/FVC value, an important indicator of lung function. (b) Relationship between smoking status and predicted risk of COPD. The figure illustrates how the risk value predicted by DeepSpiro correlates with smoking status, indicating a higher risk value is associated with a larger proportion of smokers. This result is consistent with clinical findings, as smokers are more likely to develop COPD. (c) Relationship between age and predicted risk of COPD. The figure demonstrates that the predicted risk value of DeepSpiro increases with age, indicating that older individuals have a higher risk of developing COPD. This observation is consistent with clinical findings, which show that older patients are more susceptible to COPD.}
\label{merged3}
\end{figure*}

As shown in the beeswarm plot in Figure \ref{merged3}-(a), the relative importance of features is revealed. The figure shows that the higher the COPD risk value, the greater its impact on the model. Smoking, being male, and older age all influence the model's judgment. DeepSpiro's findings are consistent with related research conclusions, indicating that smokers, males, and older patients have a higher risk of being predicted as having COPD.

We examine the relationship between smoking and risk values as well as age and risk values and present our results with dependency graphs. Figure \ref{merged3}-(b) shows the relationship between smoking and the risk of COPD. From the figure, it is evident that the predicted risk value of DeepSpiro is closely related to smoking status. The higher the risk value, the larger the proportion of smokers. This aligns with clinical significance. For COPD patients, smokers are more susceptible to the disease.

Figure \ref{merged3}-(c) illustrates the relationship between age and the risk of COPD. From the figure, it can be seen that the predicted risk value of DeepSpiro is closely related to age. The higher the risk value, the older the age. This aligns with clinical significance. For COPD patients, older patients are more susceptible to the disease.

\begin{figure*}[]
\centerline{\includegraphics[width=1\textwidth]{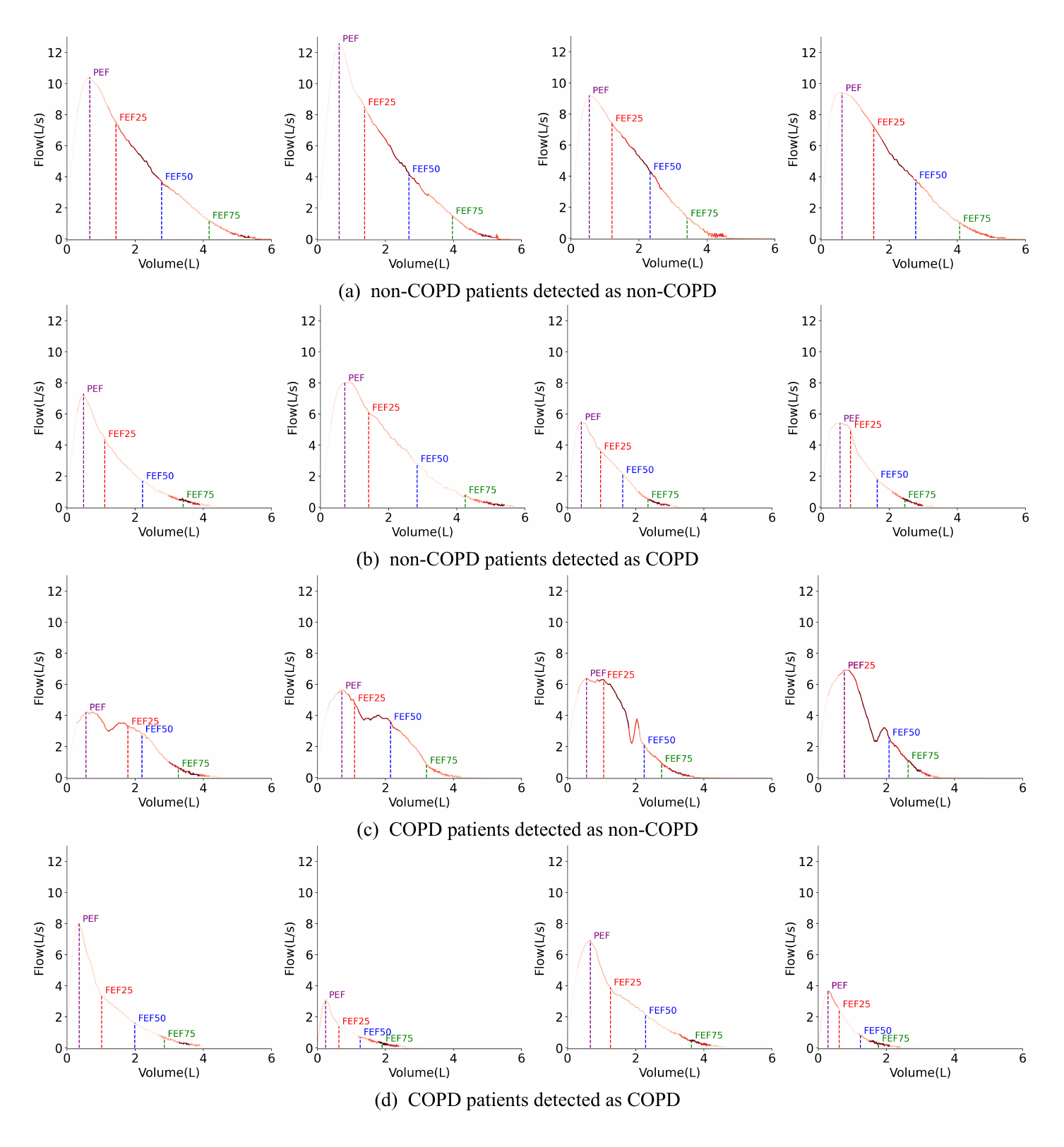}}
\caption{The explainer of the model. The brighter the color, the more attention the model pays.}
\label{attention-fig}
\end{figure*}

\subsection{Model Explainer Analysis}
Healthcare professionals often struggle to accurately identify and locate disease information directly from long sequence data in COPD classification tasks. To address this issue, we utilize SpiroExplainer, automatically focusing on the diseased regions of the Volume-Flow data.

The proposed method can automatically capture anomalies from long sequence Volume-Flow data, providing specific patches with abnormalities. Figure \ref{attention-fig} shows that the model can automatically focus on key patches that distinguish COPD patients. Brighter colors indicate higher attention in the corresponding areas, while darker colors indicate lower attention.

An interesting phenomenon is that for non-COPD patients, as shown in Figure \ref{attention-fig}-(a), the Volume-Flow curve in the $FEF_{25}$-$FEF_{50}$ area appears "full," without the small airway collapse phenomenon near $FEF_{75}$ at the tail of the curve. DeepSpiro primarily focuses on these two areas near $FEF_{25}$-$FEF_{50}$ and $FEF_{75}$, which match the characteristics of the non-COPD patient group. On the other hand, for COPD patients, as shown in Figure \ref{attention-fig}-(d), the curve's tail near the $FEF_{75}$ area collapses due to small airway collapse resulting from COPD. DeepSpiro focuses on the area near $FEF_{75}$, aligning with the characteristics of the COPD patient group. Thus, DeepSpiro can effectively differentiate and assess the features of different populations.

In analyzing cases where the model made incorrect predictions, we found that for some individuals who might have asthma, as shown in Figure \ref{attention-fig}-(b), due to insufficient inhalation, the curve after $FEF_{75}$ shows a weak exhalation characteristic similar to the small airway collapse seen in COPD patients. Since the model mainly focuses on the area near $FEF_{75}$, it leads to lower prediction accuracy for these asthma patients. For some individuals who might have COPD, as shown in Figure \ref{attention-fig}-(c), due to uneven exhalation effort or mid-breath inhalation by COPD patients, the Volume-Flow curve might present double or multiple peaks, appearing "full" in the $FEF_{25}$-$FEF_{50}$ range similar to non-COPD patients. DeepSpiro primarily focuses on areas near $FEF_{25}$-$FEF_{50}$ and $FEF_{75}$, which also leads to lower prediction accuracy for these COPD patients. We believe the model's predictions are medically interpretable, and due to the complexity and overlap of disease characteristics, it is challenging to avoid prediction errors completely.

\section{Discussion}
Recent research on COPD has primarily focused on three areas: detection, prediction, and genetic studies. The application of machine learning and deep learning has achieved significant progress in disease detection. Researchers have used advanced deep learning models, such as convolutional neural networks (CNNs), to analyze various types of data like e-nose signals \cite{avian2022fx}, spirometry data \cite{cosentino2023inference}, lung sounds \cite{dianat2023classification}, and CT images \cite{almeida2024prediction}, playing a vital role in COPD detection and classification. To improve diagnostic accuracy, efforts have also been made to integrate multi-source data, such as combining multi-omics data like proteomics and transcriptomics with disease-specific protein/gene interaction information \cite{zhuang2023deep}, or integrating imaging data with questionnaire data \cite{zhu2024development}, providing a more comprehensive diagnostic foundation. In data processing, specialized preprocessing workflows have been developed to eliminate noise and enhance dataset quality, such as denoising and feature extraction for lung sounds \cite{wu2023deep}. In the field of prediction, machine learning models have been employed to forecast COPD disease progression and survival rates \cite{yin2024machine, willer2021x}, including the use of deep neural networks to predict COPD stages \cite{yin2023fractional}, and applying anomaly detection methods to identify COPD manifestations in chest CT scans \cite{kumar2024novel}, which have shown remarkable effectiveness in detecting lung function impairment and disease severity. In genetic research, bioinformatics approaches have been used to identify key genes and genetic markers associated with COPD \cite{sharma2021determining, han2022white, yang2023revealing, lee2023meta}, deepening our understanding of COPD pathogenesis and providing valuable insights for early detection and potential therapeutic target development.

In this paper, we designed a deep learning method for detecting and early predicting COPD from the Volume-Flow curve time series. Specifically, we use SpiroSmoother, thereby enhancing the stability of the Volume-Flow curve. We propose SpiroEncoder, which unifies the temporal representation of "key patches" and achieves the conversion of key physiological information from high to low dimensions. Utilizing SpiroExplainer, we fuse diagnostic probabilities with demographic information to improve the accuracy of COPD risk assessment and the model's explainer. The results and explainer of the model can provide patients with timely COPD risk reports. Moreover, we propose SpiroPredictor, by incorporating key information such as the degree of concavity, the model can accurately predict the future disease probability of high-risk patients who have not yet been diagnosed, thereby enabling early intervention and treatment to slow the progression of the disease. 
Our proposed DeepSpiro model has two primary functions. First, it can detect COPD with high performance, achieving AUC and other performance metrics that surpass those of current SOTA models. Second, our model can predict the risk of developing COPD over the next 1-5 years using key patch concavity information, an innovative method we have proposed for the first time. Experimental results show that our novel method has high accuracy in predicting future COPD risk.

Our work does have some limitations. First, although DeepSpiro has shown good performance in detecting and predicting COPD using spirogram time series in the UK Biobank, its generalizability to other population groups remains uncertain. The characteristics of the spirogram may vary among different populations, which could potentially affect the model’s performance.

Second, while DeepSpiro demonstrates high accuracy in predicting COPD using spirogram time series in a research setting, further validation in real clinical environments is crucial. The transition from research to practical application may present unforeseen challenges, such as variations in data quality, increased noise, and differences in clinical workflows. These factors could impact the model's performance, necessitating additional adjustments and optimizations in future work.

Third, although our model demonstrated high AUROC values across most prediction intervals, its performance in terms of AUPRC was less satisfactory, particularly for predicting COPD onset within 1–2 years. We attribute this to the inherent class imbalance in the dataset (see Supplementary Table 1), where the proportion of new COPD cases is extremely low. Since AUPRC is highly sensitive to class imbalance, it may underestimate the true performance of the model in such cases. 
To investigate whether this issue stemmed from our methodology or the inherent challenges of the task, we compared our model with the approach proposed by Cosentino et al. \cite{cosentino2023inference}. The comparison results (see Supplementary Table 2 and Supplementary Figure 1) revealed that under similar experimental conditions, their model also exhibited low AUPRC values, highlighting the general difficulty of identifying rare COPD onset events in highly imbalanced datasets. Nevertheless, we observed that for certain prediction windows, such as those exceeding five years, our method achieved comparable or superior AUPRC values. This indicates that incorporating concavity features from the volume-flow curve alongside demographic information helps capture critical features over longer time horizons. 
From a clinical perspective, the importance of early COPD detection often outweighs the risk of false positives. Individuals flagged as high-risk can undergo further confirmatory assessments, such as chest imaging. Additionally, the robust AUROC performance of our model across various prediction windows demonstrates its strong discriminative capability, making DeepSpiro a valuable early screening tool. This, in turn, has the potential to improve patient care quality and optimize resource allocation in real-world medical practice.

In the future, 
we will focus on training and validating the model using datasets from diverse geographic locations and population backgrounds. This will help assess the impact of demographic and environmental factors on model performance and improve its generalizability. Additionally, 
we plan to develop user-friendly software tools that allow doctors to conveniently apply the model in clinical settings and obtain reliable and understandable results to assist in diagnosing and treating patients. Finally, we aim to improve the model's generalizability by incorporating more high-quality datasets, enabling it to provide effective predictions in a wider range of scenarios.

\section{Methods}
\subsection{Problem Definition}
For a COPD dataset $D=\{X,Y\}$ composed of $N$ instances, where $X=\{x_1,x_2,...,x_j,...,x_N\}$ represents the monitoring data for $N$ instances, and $Y=\{y_1,y_2,...,y_j,...,y_N\}$ represents the COPD disease labels for the $N$ instances. The disease label for the $j^{th}$ instance, $y_j \in \{0,1\}$, with 0 indicating no disease and 1 indicating the presence of disease. The monitoring data $x_j=\{s_j,d_j,a_j\}$ includes the spirogram data $s_j$, demographic information $d_j=\{dg_j, da_j, ds_j, ...\}$, and key concavity information $a_j=\{a_{j,pef-fef25}, a_{j,fef25-fef50},a_{j,fef50-fef75},\\a_{j,fef75+}\}$. Here, the spirogram data is a varied-length sequence $s_j=\{s_{j,1}, s_{j,2},...s_{j,i},...\}$ representing airflow variation over time during exhalation. In the demographic information, $dg_j$ denotes sex, $da_j$ represents age, and $ds_j$ indicates smoking status. The key patch concavity information includes $a_{j,pef-fef25}$ for the concavity information between PEF and FEF25, $a_{j,fef25-fef50}$ for the concavity information between FEF25 and FEF50, $a_{j,fef50-fef75}$ for the concavity information between FEF50 and FEF75, and $a_{j,fef75+}$ for the concavity information beyond FEF75.

DeepSpiro takes the monitoring data $X$ of $N$ instances as input and produces predicted disease labels $\hat{Y}=\{\hat{y}_1, \hat{y}_2,...,\hat{y}_j,..., \hat{y}_N\}$, explainer $\hat{E}=\{\hat{e}_1, \hat{e}_2,...,\hat{e}_j,..., \hat{e}_N\}$, COPD risk values $R$ and the risk coefficients contributed by each monitoring input dimension $R_s$, $R_dg$, $R_da$, $R_ds$, etc., as well as the probabilities of future disease occurrence for high-risk undiagnosed COPD patients $F_{copd}$, $F_{1year}$, $F_{2year}$, $F_{3year}$, $F_{4year}$, $F_{5year+}$, $F_{non-copd}$. Here, $y_j \in \{0,1\}$, where 0 indicates no disease, and 1 indicates the presence of disease.

\subsection{Overview}\label{overview}
\begin{figure*}[]
\centerline{\includegraphics[width=1\textwidth]{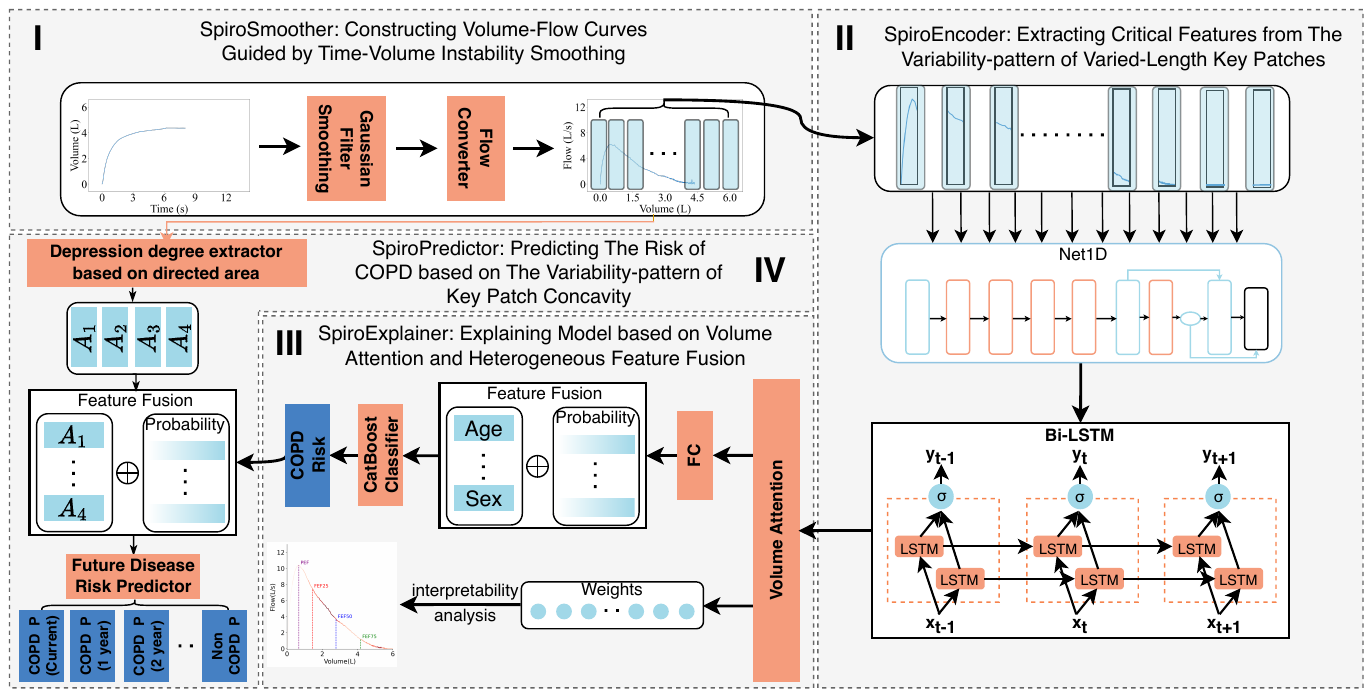}}
\caption{Framework overview. It is divided into four modules. In the first module, we process the varied-length original volume curve and convert it into a smoothed Volume-Flow curve. In the second module, we extract features from the varied-length Volume-Flow curve. In the third module, combined with demographic information, we output the COPD detection results and model explainer. In the fourth module, based on concavity information, we output the risk values for COPD at different future periods.}
\label{framwork-overview-fig}
\end{figure*}

Figure \ref{framwork-overview-fig} illustrates the architecture of DeepSpiro, where the blocks represent different components of the framework. DeepSpiro is primarily divided into two tasks: COPD detection and early COPD prediction. The inputs to DeepSpiro always include spirometry data and demographic information (such as age, gender, smoking status, etc.). For the COPD detection task, the model outputs interpretability data and a COPD risk score. For the early prediction task, it outputs the probability of the individual developing COPD within the next 1-5 years. It is important to note that DeepSpiro only proceeds with the early prediction task if the detection task determines that the individual does not currently have COPD. Therefore, if the detection results indicate that the patient has COPD, the model’s final output is whether the patient has COPD. If not, the output includes both the current COPD status and the probability of developing COPD in the next 1-5 years.

DeepSpiro can be divided into four main modules: 1) SpiroSmoother (see Figure \ref{framwork-overview-fig}-I): constructing Volume-Flow curves guided by Time-Volume instability smoothing: This enhances the stability of key physiological information in the original Volume-Flow data. 2) SpiroEncoder (see Figure \ref{framwork-overview-fig}-II): extracting critical features from the variability-pattern of varied-length key patches: This dynamically identifies "key patches" and unifies the time-series representation, achieving the conversion of key physiological information from high-dimensional to low-dimensional. 3) SpiroExplainer (see Figure \ref{framwork-overview-fig}-III): explaining model based on volume attention and heterogeneous feature fusion: This combines diagnostic probabilities with demographic information to assess COPD risk and provides an explainer of the model's decisions. 4) SpiroPredictor (see Figure \ref{framwork-overview-fig}-IV): predicting the risk of COPD based on the variability-pattern of key patch concavity: This integrates key concavity information to assess COPD risk for various future periods.

Specifically, the original spirogram data is processed with signal enhancement technology to obtain smoothed Volume-Flow data. The smoothed Volume-Flow data is used as input in DeepSpiro for varied-length time series data in order to extract essential physiological information. Subsequently, the model combines the important physiological details and relevant demographic information using volume attention and heterogeneous feature fusion techniques. This process enables the model to output a COPD risk assessment and provide an explainer for its decisions. By using the COPD risk assessment values and concavity information obtained from the Volume-Flow signals in the feature fusion model, we can make precise predictions about the probability of future disease in high-risk patients who have not yet received a diagnosis.

\subsection{SpiroSmoother: Constructing Volume-Flow Curves Guided by Time-Volume Instability Smoothing}
\begin{figure*}[]
\centerline{\includegraphics[width=1\textwidth]{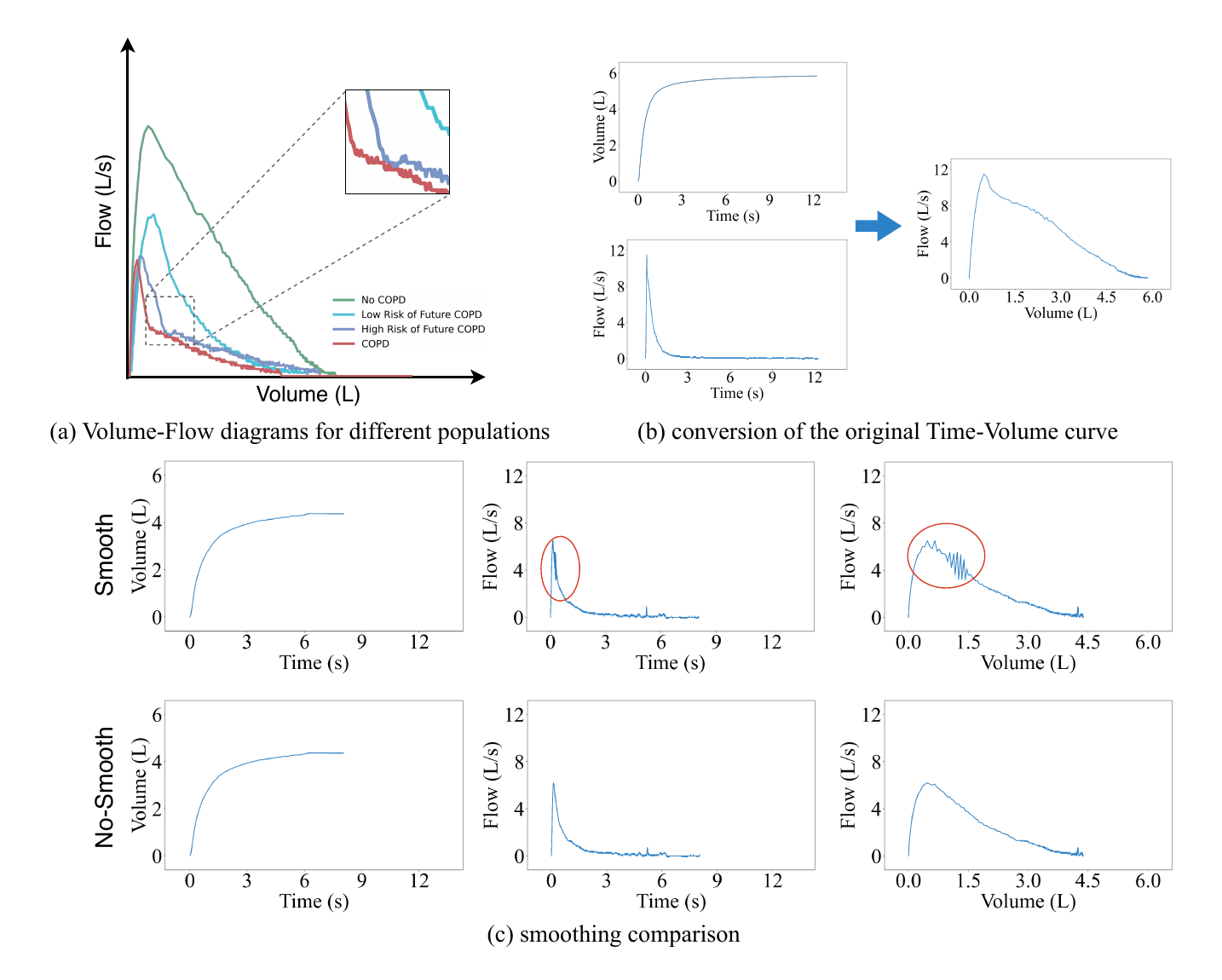}}
\caption{(a) Examples of Volume-Flow diagrams for different populations. (b) The conversion of the original Time-Volume curve. To obtain the degree of airflow limitation, we need to use finite difference methods on the original Time-Volume curve to calculate the corresponding Volume-Flow curve. (c) Smoothing comparison, the original curve shows fluctuations in some areas when converting to Time-Flow and Volume-Flow curves. After smoothing, the curves in the same positions become more stable.}
\label{spirogram-introduce}
\end{figure*}

The current state-of-the-art (SOTA) method for detecting COPD from the Volume-Flow curve using deep learning has been proposed by J. Cosentino et al. \cite{cosentino2023inference}. This method involves converting original Time-Volume curves into Volume-Flow curves, which can result in instability in the curves (As shown in Figure \ref{spirogram-introduce}-C). Extensive experimental validation has been demonstrated that these instabilities can impact the final risk assessment for COPD patients. To address this, we employ SpiroSmoother. This approach precisely enhances the stability of the original Volume-Flow curves while preserving key physiological signals within the original Volume-Flow data.

As shown in Figure \ref{spirogram-introduce}-C, we observe instability when converting the Time-Volume curve into a Time-Flow curve. Therefore, we believe that smoothing the original Time-Volume curve could help enhance the stability of both the Time-Flow curve and the Volume-Flow curve.

Gaussian filtering is employed to implement instability smoothing guidance for the original Time-Volume curves. This method achieves smoothing by taking a weighted average of the areas around the data points, where the weight of each data point is determined by a Gaussian function. For a given data point $x_i$ its smoothed value $y_i$can be defined as:
\begin{equation}
y_i = \frac{\sum_{j=-k}^{k} x_{i+j} \cdot g(j)}{\sum_{j=-k}^{k} g(j)}
\end{equation}
In this formula, $g(j)$ represents the Gaussian function,$k$ is the filter's window size, and $x_{i+j}$represents the original data point and its adjacent points. The formula calculates the weighted average of data points around $x_i$, using weights derived from the Gaussian function $g(j)$. This function is centered at $x_i$, and its standard deviation determines both the range and the degree of smoothing. Through this method, Gaussian filtering effectively smooths the data while preserving important feature information.

To convert the smoothed Time-Volume curve into a Volume-Flow curve, we utilize the finite difference method to approximate the first derivative of the volume data with respect to time. Simply put, it involves calculating the volume change rate at consecutive time points to obtain corresponding flow data. The flow $Q(t)$ can be obtained by calculating the derivative of volume $V(t)$ with respect to time $t$, where flow $Q(t)$ is defined as:
\begin{equation}
Q(t) \approx \frac{V(t+\Delta t) - V(t)}{\Delta t}
\end{equation}
Where $V(t)$ represents a set of Time-Volume curve data, $\Delta t$ is the time interval between two adjacent time points, and $t$ denotes the time points.

We linearly interpolate the calculated flow data to ensure that the Time-Flow curve is consistent with the original Time-Volume curve in terms of time data. In this way, we can construct the Volume-Flow curves $\mathscr{F}(v)$ using the Time-Flow curves $Q(t)$ and the Time-Volume curves $V(t)$. The Volume-Flow curves $\mathscr{F}(v)$ is defined as:
\begin{equation}
 \mathscr{F}(v) = Q(V^{-1}(v)) 
\end{equation}

As shown in Figure \ref{spirogram-introduce}-C, the Time-Flow curve and the Volume-Flow curve have been constructed using the Time-Volume instability smoothing guidance technique and have effectively eliminated the original fluctuations, significantly enhancing the data's stability. In subsequent experiments, we have used these stable Volume-Flow curves to replace the original data for model training. Experimental results have confirmed that the curves processed through instability smoothing guidance demonstrate superior performance in risk assessment.

\subsection{SpiroEncoder: Extracting Critical Features from The Variability-pattern of Varied-Length Key Patches}
Existing methods for modeling varied-length time series data generally use missing value imputation or downsampling truncation. The former introduces external noise, while the latter will likely lose meaningful data dependencies. We aim to preserve the key signals of the original data and adapt them to model input. To this end, we propose SpiroEncoder.

First, we need to reconstruct the original data. We have adopted an adaptive temporal decomposition method, which dynamically calculates the most suitable "key patches" count for each time series patch, accurately capturing the key Volume-Flow information within each time patch. By introducing the hyperparameter $k$, we can calculate the number of key patches based on the ratio of the sequence length to $k$, ensuring that each patch contains enough information for subsequent analysis. For a given length of time series $L$, the number of key patches $S$ can be defined as:
\begin{equation}
S = \left\lceil\frac{L}{k} \right\rceil
\end{equation}
Where $\lceil \cdot \rceil$ represents the ceiling function, ensuring that meaningful data dependencies are not lost even when $L$ is not divisible by $k$.

We can obtain the number of key patches for each sequence using the abovementioned method. Thus, we can divide each sequence into several key patches. We feed all the key patches into the Net1D network \cite{hong2020holmes} to extract the spatial features between the patches and the temporal features within each patch. The Net1D network extracts spatial features for us. We must compress the high-dimensional spirogram space into the same low-dimensional key patch to adapt these features for our subsequent modules.

First, we determine the length of the most extended sequence in the dataset and use this length along with the preset hyperparameter $k$. Calculate the number of key patches all sequences should have, thereby normalizing the temporal representation of the data. For the longest sequence length $L_{\text{max}}$ in the dataset, the maximum number of patches is defined as:
\begin{equation}
N_{\text{max}} = \left\lceil \frac{L_{\text{max}}}{k} \right\rceil
\end{equation}
We perform encoding for the key features of each sequence. This encoding marks the key patches and records them as a mask. For each sequence, we create a mask vector. Parts of the vector where key patches exist are masked as 1, and parts extending from the end of the existing key patches to the maximum number of patches are masked as 0.

We construct a zero tensor $O \in \mathbb{R}^{N \times S \times C}$, where $N$ represents the number of samples, $S$ represents the number of key patches, and $C$ represents the number of output channels. Subsequently, we selectively apply these encodings to the zero tensor $O$ using the mask vector. For each sample, the mask vector indicates where to insert the features of key patches. Specifically, if the mask vector is one at a certain position, then we insert the encoding at the corresponding position in the zero tensor $O$; if the mask vector is 0 at a position, the zero tensor remains unprocessed at that position (This part also constitutes the padding of 0s for the sequence). Thus, the zero tensor $O$ transforms into a fixed-length sequence containing encoding information, and we refer to the transformed zero tensor $O$ as the feature tensor of key patches $O$. For each sample $i$ and its key patches $j$, we use the mask vector $M_{i,j}$ to decide whether to apply encoding. $O_{i,j,:}$ can be defined as:
\begin{equation}
O_{i,j,:} = M_{i,j} \cdot E_{i,j,:}
\end{equation}
where the features extracted by Net1D for each key patch are $E_{i,j,:}\in \mathbb{R}^C$

Because the process of constructing the key patch feature tensor $O$ involves padding with zeros, this padded information may introduce unnecessary noise when captured by the Bidirectional Long Short-Term Memory Network (Bi-LSTM). To address this, we aim to effectively ignore this portion of padded information before entering the Long Short-Term Memory Network.

First, we calculate the effective length $Length_i$ for each sequence, which is the number of valid key patches in the sequence $i$. The effective length $Length_i$ can be defined as:

\begin{equation}
Length_i = \sum_{j=1}^{S} M_{i,j}
\end{equation}

Then, we create a new tensor $P \in \mathbb{R}^{T \times C}$, where $T = \sum_{i=1}^{N} Length_i$, that is, the sum of the effective lengths of all sequences, and $C$ is the number of output channels. Following the sequence order, we copy the features of valid key patches from the key patch feature tensor $O$ into the tensor $P$. After the above steps, the tensor $P$ contains only valid key patch features.

After the tensor $P$ enters the Bidirectional Long Short-Term Memory network (Bi-LSTM) \cite{hochreiter1997long}, the variability-pattern relationships among key patches is obtained.

\subsection{SpiroExplainer: Explaining Model based on Volume Attention and Heterogeneous Feature Fusion}
Existing deep learning models still function as black boxes, capable only of producing diagnostic outcomes without offering an explainer for those results. In order to establish credibility with medical professionals and patients, the model must offer a decision-making process that is more open and easily understood by both parties. 
Therefore, through the dynamic volume attention integration method, we accurately highlight the time patches crucial for model predictions using a method for explaining model based on volume attention and heterogeneous feature fusion. 

Data enters the first linear transformation layer as it passes through the volume attention layer. This layer provides an initial feature representation for each time step, which is specifically manifested as:
\begin{equation}
X' = W_1X + b_1
\end{equation}
Here, $W_1$ and $b_1$ are the weight and bias of the linear layer, respectively.

To perform a nonlinear transformation, We follow up by applying the Swish activation function to the linearly transformed data $X'$.
The Swish activation function helps to improve the model’s ability to capture complex features. The transformed data after applying Swish activation function is given by:
\begin{equation}
X'' = X' \cdot \sigma(X')
\end{equation}
where $\sigma(X') = \frac{1}{1 + e^{-X'}}$ is the sigmoid function applied element-wise to $X'$.
Next,
we apply a bilinear transformation to the activated data $X"$ by using a bilinear weight matrix. This is specifically represented as:
\begin{equation}
X''' = X''W_{\text{bil}}
\end{equation}
where $W_{\text{bil}}$ is the bilinear weight matrix.

After passing through the bilinear transformation, the data $X'''$ goes through another linear layer to calculate the attention scores $S$ for each time step. Finally, the $S$ scores are normalized using the $softmax$ function to obtain the volume attention weights.

We combine the volume attention information with the smoothed Time-Volume curve data. The decision-making process within the deep neural network model can be transformed into visualized graphs. The graphs provide a clear representation of the specific data regions that the model prioritizes when making decisions, thereby improving the transparency of the model and bolstering the credibility of its decision-making process.

Research indicates that there is a significant correlation between smoking, age, and other demographic information with COPD. Therefore, we aim to incorporate some demographic information to enhance the effectiveness of our model. For this purpose, after passing through the volume attention layer, we first conduct an initial COPD assessment through a fully connected layer, resulting in $\hat{P}=\{\hat{p}^1, \hat{p}^2,...,\hat{p}^j,..., \hat{p}^N\}$, where $\hat{P}$ represents a series of COPD assessment values for the samples, with $\hat{p}^j \in [0,1]$ denoting the assessment value for the $j$-th sample. 
In the subsequent steps, we use these evaluation values to generate the feature $F_{\text{probability}}$, representing the probability of disease. Specifically, $F_{\text{probability}}$ is obtained by applying the $Softmax$ function to $\hat{P}$, as follows:
\begin{equation}
F_{\text{probability}} = \text{Softmax}(\hat{P})
\end{equation}
Additionally, we extract features $F_{\text{struct}}$ from demographic and other structured data. Then, we combine these features with $F_{\text{probability}}$ using the following fusion operation:
\begin{equation}
F_{\text{fusion}} = F_{\text{probability}} \oplus F_{\text{struct}}
\end{equation}
where $\oplus$ denotes the concatenation of features.

By utilizing the Gradient Boosting framework (Catboost) to process the fused features $F_{\text{fusion}}$, we predict the risk of COPD. The COPD risk assessment obtained in this manner outperforms the assessment results derived solely from unstructured varied-length Volume-Flow curve data.

\begin{figure*}[]
\centerline{\includegraphics[width=1\textwidth]{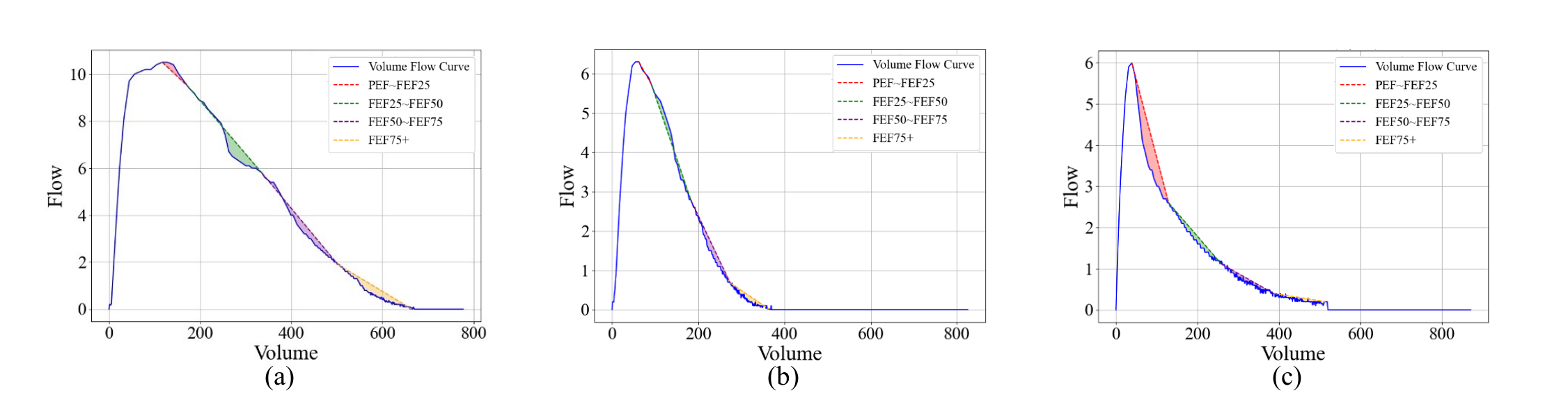}}
\caption{Volume-Flow curves illustrating phase-specific concavity and airway collapse. (a) A representative Volume-Flow curve divided into four phases: early (PEF$\sim$FEF25), mid-early (FEF25$\sim$FEF50), mid-late (FEF50$\sim$FEF75), and late (FEF75+), highlighting concave and convex segments relative to the baseline. (b) Volume-Flow curve of a healthy individual, showing airway collapse occurring in the late phase (FEF75+). (c) Volume-Flow curve of a COPD patient, showing early-stage airway collapse in the early phase (PEF$\sim$FEF25).}
\label{volume-flow-curve}
\end{figure*}

\subsection{SpiroPredictor: Predicting The Risk of COPD based on The Variability-pattern of Key Patch Concavity}
In future risk prediction tasks, we used the degree of concavity in individual Volume-Flow graphs as one of the features for prediction. As shown in Figure \ref{spirogram-introduce}-A, high-risk individuals and COPD patients exhibit more significant concavity, while low-risk individuals exhibit fewer concavity than high-risk individuals and COPD patients. The following provides a detailed explanation of how this feature of concavity is extracted.

Since the spirogram itself is not strictly monotonic, each phase consists of a set of concave and convex curve segments (such as the FEF50$\sim$FEF75 phase in Figure \ref{volume-flow-curve}-A), making it difficult to simply judge the degree of collapse in the curve. Therefore, we have defined a concavity-based directed area metric to approximate the degree of collapse in such non-monotonic curves, in order to characterize the overall concave-convex nature exhibited by the curve.

First, given the Volume-Flow curve $\mathscr{F}(v)$ for a particular phase $s$, we define its baseline $BL(s, v)$ as the straight line connecting the starting and ending points of the curve in that phase, where the starting volume of the phase curve is $B(s)$ and the ending volume is $G(s)$. The slope of the baseline $BL(s, v)$, denoted as $m(s)$, can be defined as:
\begin{equation}
m(s) = \frac{\mathscr{F}((G(s)) - \mathscr{F}( (B(s))}{G(s) - B(s)}
\end{equation}
The intercept $b(s)$ of the baseline $BL(s)$ can be defined as:
\begin{equation}
b(s) = \mathscr{F} (B(s)) - m(s) \times B(s)
\end{equation}
Thus, the baseline $BL$ for a given phase $s$ can be expressed as:
Where $v$ is a specific volume value on the Volume-Flow curve.
Next, we define a concavity measure $\mathscr{C}(s)$ based on directed area, which is the directed area of phase $s$:
\begin{equation}
\mathscr{C}(s) = \sum_{v=B(s)}^{G(s)} (BL(s, v) - \mathscr{F}(v))
\end{equation}

Therefore, if the curve of phase $s$ is convex, meaning that the Volume-Flow curve of this phase lies above its baseline (as in the PEF$\sim$FEF25 phase in Figure \ref{volume-flow-curve}-A), the directed area is negative, i.e., the concavity measure is negative. Conversely, if the curve of phase $s$ is concave, meaning that the Volume-Flow curve of this phase lies below its baseline (as in the FEF25$\sim$ FEF50 phase in Figure \ref{volume-flow-curve}-A), the directed area is positive, i.e., the concavity measure is positive. According to the above definition, the more concave a curve is, the larger the concavity measure, and conversely, the more convex a curve is, the smaller the concavity measure. The concavity measure of a given phase represents the concavity information for that phase.

Furthermore, for Volume-Flow curves in some phases that are neither strictly monotonic increasing nor monotonic decreasing, the total concavity measure of phase s depends on the sum of the concave area (positive value) and the convex area (negative value). If the concave area in phase $s$ is larger, the concavity measure is greater (positive value), whereas if the convex area is larger, the concavity measure is smaller (negative value). For example, in the FEF50$\sim$FEF75 phase in Figure \ref{volume-flow-curve}-A, the concave area is much larger than the convex area. Thus, by calculation, the concavity measures $\mathscr{C}(FEF50\sim FEF75)$ based on directed area is positive, indicating that this phase exhibits a significant concave trend, rather than a convex one.

If we divide the spirogram into four phases: early (PEF$\sim$FEF25), mid-early (FEF25$\sim$FEF50), mid-late (FEF50$\sim$FEF75), and late (FEF75+), we find that in healthy individuals, the spirogram typically remains full during the early phases, and collapse does not occur until the late phase (such as in the FEF75+ phase, as shown in Figure \ref{volume-flow-curve}-B). In contrast, COPD patients often experience early-stage collapse (such as in the PEF$\sim$FEF25 phase, as shown in Figure \ref{volume-flow-curve}-C). Therefore, we aim to use the proposed concavity measure based on directed area to study the relationship between the phase in which collapse occurs and future disease risk.

We fused the COPD Risk output from the SpiroExplainer module with the key concavity information of the Volume-Flow curve and input this combined data into a future disease risk predictor for assessing future risk. Our future disease risk predictor can be classifiers like Catboost \cite{wei2023risk}, Xgboost \cite{raihan2023detection}, Random Forest \cite{sun2024improved}, etc. The model, having been obtained in this way, accurately predicts the probability of illness in the next 1, 2, 3, 4, 5 years, and beyond for high-risk patients who have not yet been diagnosed.

\subsection{Dataset and Preprocessing}
The data used in our study was obtained from the UK Biobank. We specifically focused on information from 453,558 patients who underwent their initial pulmonary function test (The test outputs the Time-Volume time-series curve). 
It is worth noting that due to factors such as genetics, environment, or lifestyle, the spirogram characteristics may vary among different population groups. Therefore, in our data processing, we chose to conduct experiments only on the largest population group in the UK Biobank, which is of European descent.
The DeepSpiro model relies on detailed and precise patterns in spirogram time series to detect and predict COPD. Therefore, incomplete or distorted data may lead to errors in the model's predictions. To mitigate the issue of inaccurate predictions caused by poor data quality, we have implemented several data preprocessing measures. These measures ensure that only high-quality and reliable data is used for model training and validation, thereby enhancing the robustness and accuracy of the DeepSpiro model.

The original Time-Volume (Expiratory volumes over a period of time) measurements are extracted from UK Biobank field 3066, which contains expiratory volumes recorded every 10 milliseconds in milliliters. The Time-Volume measurements are used from each participant's first visit. To ensure the Time-Volume data are valid, we consult UK Biobank field 3061; if the value in field 3061 is either 0 or 32, the expiration is considered valid. If multiple valid expirations are available, we choose the first one in sequence. To control the quality of the expirations, we review the UK Biobank fields 3062 (FVC), 3063 (FEV1), and 3064 (PEF); if any of these are in the top or bottom 0.5\% of the observed values, any expiratory data will discard from that patient.
The original expiratory volume measurements are converted from milliliters to liters, and the corresponding flow curves are calculated by approximating the first derivative with respect to time using finite differences. As shown in Figure \ref{spirogram-introduce}-B, the Time-Flow and Time-Volume curves combine to generate a one-dimensional Volume-Flow curve time series.

We refer to the labeling methods in Justin Cosentino et al.'s research \cite{cosentino2023inference}. Specifically, the labels are generated by combining information from various sources in the UK Biobank. We use a binary COPD label that is determined through self-reporting, hospital admissions, and primary care reports to train DeepSpiro.
Self-reported COPD status is derived from codes 1112, 1113, and 1472 in the UK Biobank field 20002. Hospital-reported COPD status comes from codes like J430 and others in the UK Biobank field 41270, and codes like 4920 and others in field 41271. The COPD status from primary care reports is required using TRUD to map the UK Biobank's gp-clinical to read codes from UK Biobank field 42040, with mapped fields including codes like J430, among others. For specific extraction of field codes, please refer to the Table \ref{label table}.
In addition, death information is used in subsequent analyses. Death information is recorded in UK Biobank field 40000. If any value is present in this field, we consider the patient to have died.

\begin{table}[ht]
\centering
\caption{The following table lists the required fields and corresponding codes during label extraction.}
\label{label table}
\begin{tabularx}{\columnwidth}{lXl}
\toprule
Field Id & Code & Type \\
\midrule
20002  & 1112, 1113, 1472 & Self Report \\
\cmidrule(lr){1-3}
\multirow{3}{*}{41270} 
       & J430, J431, J432, J438   & \multirow{3}{*}{Hospitalization} \\
       & 439J, J440, J441, J448  & \\
       & J449                    & \\
\cmidrule(lr){1-3}
41271  & 4920, 4928, 4929, 496X   & Hospitalization \\
\cmidrule(lr){1-3}
\multirow{3}{*}{42040}
       & J430, J431, J432, J438   & \multirow{3}{*}{Primary Care} \\
       & 439J, J440, J441, J448  & \\
       & J449                    & \\
\bottomrule
\end{tabularx}
\footnotesize{Note: A code suffixed with ‘X’ stands for any code starting with the figures preceding the X.}
\end{table}

After completing the necessary data processing, the data were categorized into three groups: All (the complete dataset), Hospitalization (patients reporting hospitalization), and Death (patients who passed away). The number of COPD cases in each category is summarized in the table below:

\begin{table}[ht]
\centering
\caption{The number of COPD cases under different categories.}
\label{label table}
\begin{tabularx}{\columnwidth}{lXX}
\toprule
Category & Nums & COPD Case \\
\midrule
All              & 348039        & 19308       \\
Hospitalization & 348039        & 16254       \\
Death           & 24105         & 4243        \\
\bottomrule
\end{tabularx}
\end{table}

Currently, the dataset includes a total of 348,039 participants, divided into training and testing sets at an 8:2 ratio. Specifically, the training set contains 278,431 participants, while the testing set comprises 69,608 participants.

\subsection{Evaluations}
To evaluate the performance of this method, we used a validation set to calculate the model's AUROC (Area Under the Receiver Operating Characteristic curve), AUPRC (Area Under the Precision-Recall Curve), and F1 scores.

AUROC: The area under the receiver operating characteristic curve (AUROC) is a widely used metric for evaluating the performance of classification models, especially in binary classification problems. It is calculated by varying the classification threshold to determine the true positive rate ($TPR = \frac{TP}{TP + FN}$) and the false positive rate ($FPR = \frac{FP}{FP + TN}$). The ROC curve shows the relationship between the true positive rate and the false positive rate. The value of AUROC is the area under the ROC curve, which can be approximated as $AUROC = \int_{0}^{1} R(F) \, dF$, where $F$ represents the false positive rate and $R(F)$ is the corresponding true positive rate. Numerical methods generally estimate the FPR values $f_1, f_2, \ldots, f_n$ and the corresponding TPR values $t_1, t_2, \ldots, t_n$. The trapezoidal area approximation of the AUROC is: $AUROC \approx \sum_{i=1}^{n-1} \frac{(f_{i+1} - f_i) \times (t_i + t_{i+1})}{2}$.

AUPRC: The area under the precision-recall curve (AUPRC) is a commonly used metric to evaluate models for classification problems. The PR curve is derived by varying the classification threshold to calculate precision ($precision = \frac{TP}{TP + FP}$) and recall ($recall = TPR = \frac{TP}{TP + FN}$). It can be approximated as $\int P(r) \, dR(r)$, where $P(r)$ represents precision, $R(r)$ denotes recall, and $r$ represents the decision threshold.

F1 Score: The F1 score is a crucial metric for evaluating the accuracy of a model. It is the harmonic mean of precision and recall ($F1 = 2 \times \frac{\text{precision} \times \text{recall}}{\text{precision} + \text{recall}}$), providing a single metric that considers both the accuracy and completeness of the model.

These metrics collectively help us comprehensively assess the model's performance across different aspects. In this experiment, the AUROC, AUPRC, and F1 scores are calculated using functions from the scikit-learn library and code from Justin Cosentino et al. \cite{cosentino2023inference}.

\subsection{Compared Methods}
To assess the effectiveness of DeepSpiro, we compared it against several baseline methods, including:
\begin{itemize}
\item FEV1/FVC ratio: The FEV1/FVC ratio is the GOLD standard in clinical practice for determining whether an individual has COPD. If an individual's ratio is less than 70\%, they are considered to have COPD. Therefore, this method also be used as our baseline.

\item ResNet18 (Nature Genetics): The ResNet18 convolutional neural network (CNN) is capable of efficiently learning features from large-scale datasets. In a paper published in Nature Genetics, Justin Cosentino et al. utilized this model for the task of COPD detection \cite{cosentino2023inference}. Therefore, we also adopt this model as the primary baseline.
\end{itemize}

\section*{Data availability}
Data from the UK Biobank, which is available after the approval of an application at \url{https://www.ukbiobank.ac.uk}. UK Biobank received ethical approval from the National Information Governance Board for Health and Social Care and the National Health Service North West Centre for Research Ethics Committee (Ref: 21/NW/0157). 

\section*{Code availability}
Our code is publicly available at \url{https://github.com/yudaleng/COPD-Early-Prediction}.

\section*{Acknowledgement}
The authors gratefully acknowledge the financial supports by the National Natural Science Foundation of China under Grant 62202332, Grant 62102008, Grant 62376197, Grant 62020106004 and Grant 92048301; Clinical Medicine Plus X - Young Scholars Project of Peking University, the Fundamental Research Funds for the Central Universities (PKU2024LCXQ030); PKU-OPPO Fund (B0202301); Beijing Natural Science Foundations (QY23040); Natural Science Foundation of Tianjin City (23JCYBJC00360); College Student Innovation and Entrepreneurship Training Program (202410060109).

\section*{Author Contributions}
SM, YZhou, and SH conceptualized the study. SM, YZhou, and SH developed the methodology. SM developed the software. SM, YZhou, SH, and SC validated the results. SM, SC, and YZhou performed the formal analysis. JXu, YW, SG, and QZ carried out the investigation. SM, SH, and JXie were responsible for data curation. SM, YZhou, and SH wrote the original draft. YZhou, YZhang, and SH reviewed and edited the manuscript. SM and YW were responsible for visualization. YZhou, YZhang, XL, SY and SH supervised the project. YZhou and SH administered the project. SM, SH, YZhou, and SY received funding support for this work. All authors read and approved the final manuscript.

\section*{Competing Interests}
The authors declare no competing interests.

\bibliographystyle{unsrt}

\bibliography{sn-bibliography}

\end{document}